\documentclass[10pt,twocolumn,letterpaper]{article}

\usepackage{iccv}              

\usepackage[accsupp]{axessibility}
\definecolor{iccvblue}{rgb}{0.21,0.49,0.74}
\usepackage[pagebackref,breaklinks,colorlinks,allcolors=iccvblue]{hyperref}

\def\paperID{*****} 
\def\confName{ICCV}
\def\confYear{2025}

\title{Better Supervised Fine-tuning for VQA: Integer-Only Loss}

\newcommand{\IEEEauthorrefmark}[1]{\textsuperscript{#1}}
\author{
		Baihong Qian\IEEEauthorrefmark{1},
		Haotian Fan\IEEEauthorrefmark{1},
		Wenjie Liao\IEEEauthorrefmark{1}, 
		Yunqiu Wang\IEEEauthorrefmark{1}, 
		Tao Li\IEEEauthorrefmark{1},
		and
            \vspace{4mm}
		Junhui Cui\IEEEauthorrefmark{1} \\
	\IEEEauthorrefmark{1}ByteDance, Xinjiangwan Square, Shanghai, China\\
}

\begin{document}
\maketitle
\begin{abstract}
With the rapid advancement of vision language models(VLM), their ability to assess visual content based on specific criteria and dimensions has become increasingly critical for applications such as video-theme consistency assessment and visual quality scoring. However, existing methods often suffer from imprecise results and inefficient loss calculation, which limit the focus of the model on key evaluation indicators. To address this, we propose IOVQA(Integer-only VQA), a novel fine-tuning approach tailored for VLMs to enhance their performance in video quality assessment tasks.
The key innovation of IOVQA lies in its label construction and its targeted loss calculation mechanism. Specifically, during dataset curation, we constrain the model’s output to integers within the range of [10,50], ensuring numerical stability, and convert decimal Overall\_MOS to integer before using them as labels. We also introduce a target-mask strategy: when computing the loss, only the first two-digit-integer of the label is unmasked, forcing the model to learn the critical components of the numerical evaluation.
After fine-tuning the Qwen2.5-VL model using the constructed dataset, experimental results demonstrate that the proposed method significantly improves the model’s accuracy and consistency in the VQA task, ranking 3rd in VQualA 2025 GenAI-Bench AIGC Video Quality Assessment Challenge -- Track I. Our work highlights the effectiveness of merely leaving integer labels during fine-tuning, providing an effective idea for optimizing VLMs in quantitative evaluation scenarios.

\end{abstract}

\section{Introduction}
\label{sec:intro}
Nowadays, there has been a rapid rise in powerful text-to-video (T2V) generative models. A variety of T2V models, such as Sora\cite{Sora}, Runway Gen-2\cite{gen2}, Lumiere\cite{Lumiere} has achieved unprecedented photorealism, demonstrating their capability of creating videos with longer durations and higher quality. However, problems like unnatural elements, inconsistencies, and hallucinations still exist, leading to a need for dependable fine-grained metrics for evaluation. \par

In recent years, a wide variety of metrics have been put forward to evaluate video quality. The earliest metrics are holistic quality ones like FVD\cite{fvd} and IS\cite{is}. Some metrics like CLIP\cite{clip} and DINO\cite{DINO} that specify on visual quality or text alignment compute image-text embedding correlations, but are ineffective in other areas such as motion smoothness and factual consistency. Metrics like T2VQA\cite{T2VQA}, FastVQA\cite{Fast-vqa} and DOVER\cite{dover} only focus on a single mean opinion score (MOS) and fail to provide detailed sub-scores across various aspects. There are also studies like VIEScore\cite{Viescore} and VideoPhy\cite{Videophy} suggest using multi-modal large-language-models (MLLM) like GPT-4o\cite{GPT4o} or Gemini-1.5\cite{Gemini} to conduct multi-aspect quality assessments of given videos, while still having low correlation with human evaluations, limiting the development of better video-generative models.\par

To address the challenges highlighted before, we propose an approach to develop a visual quality assessment model that better aligns with human perception, through fine-tuning Qwen2.5-VL of different parameter counts. The key idea is to bridge the gap between computational metrics and human subjective assessments by integrating domain-specific knowledge into data pre-processing and optimizing the model’s learning process to capture human preferences towards video quality.\par

To construct the training input, we utilize the given dataset by prompt engineering and label pre-processing. In prompt engineering, we directly embeds human-centric criteria, which is aesthetic quality, image quality, temporal quality and text-video alignment in this challenge, and their detailed explanation into the model’s prompt, guiding it to assess videos along dimensions that matter most to human judgment. So that we can overcome the limits of inherent biases in single-model assessments in previous metrics.\par

Moreover, using the given score, which are decimal numbers, may not necessarily yield ideal results. The reason lies in intrinsic properties of loss functions and model learning mechanisms.\par

To prevent the model from generating rationales for its scoring outputs, the temperature parameter must be set to 0 in the process of response generation. This setting enforces deterministic response, since the argmax function forces the model to select the token with the highest predicted probability. However, the model producing identical scores across repeated predictions will eliminate the possibility of producing more fine-grained scores when given the same number of grading level as given to human evaluators. Because the score of human assessment is an average of several ones. Consequently, the model’s single-output score becomes the sole basis for evaluation, increasing reliance on the accuracy of individual predictions.\par

Furthermore, an inherent advantage in predicting integer values over decimal numbers exist in large language models, rooted in their auto-regressive policy as categorical predictors. Auto-regressive policy operates by sequentially predicting discrete tokens from a vocabulary, with each prediction step introducing potential uncertainty. Decimal scores require an additional token for the decimal point, for example, "3.5" as three tokens: "3", "." and "5", which extend the prediction chain and amplify cumulative errors from sequential categorical decisions. In contrast, integers reduce the number of prediction steps and minimize the propagation of uncertainty. This structural advantage makes integer predictions more robust and aligned with the model’s intrinsic capability for precise scoring inference.\par

As a result, we strictly require the VLM to output an integer ranging from 10 to 50 to avoid decimal point in loss computation. 
In label pre-processing, we also convert the given Overall\_MOS, which may be decimal numbers, into integers before using as labels. \par

Fine-tuning VLMs on this preprocessed data enhances the model’s ability to mimic human perception. By training the model to predict the Overall\_MOS scaled to 10-50, which are derived from aggregated human judgments, we enable the model to learn the implicit patterns in how humans balance given dimensions and provide an overall evaluation. This stands in contrast to metrics like T2VQA\cite{T2VQA} or FastVQA\cite{Fast-vqa}, whose assessment is limited to a single MOS without capturing the multidimensional nature of human assessment. Moreover, the constraint on the model to generate numerical scores only, rather than free-text explanations, ensures consistency in evaluation results, making it comparable across different videos and scenarios, which solves the low correlation problem of existing MLLM-based methods like VIEscore\cite{Viescore} or VideoPhy\cite{Videophy}.\par

By combining label pre-processing with fine-tuning that aligns the model’s outputs with human-derived MOS, the proposed approach aims to create an evaluation model that not only covers multiple assessment dimensions but also maintains high correlation with human perceptions. Such a model could serve as a reliable benchmark for video-generative models, overcoming the current bottleneck in their development caused by misalignment between computational metrics and human evaluation standards.\par

In conclusion, our contributions are listed as follows:
\begin{itemize}
    \item We propose a novel supervised fine-tuning method, IOVQA, which converts decimal labels into integers and employs an integer-only mask in loss computation, enabling VLMs to assess video quality in a manner that aligns more closely with human perception.
    \item We add evaluation dimension into prompts and utilize structural output to boost model performance and stability.
    \item We conducted experiments on the test dataset of VQualA 2025 GenAI-Bench AIGC Video Quality Assessment Challenge -- Track I and our method ranked 3rd, proving the effectiveness of IOVQA.
\end{itemize}

\section{Related Work}
\label{sec:related}
\begin{figure*}[tbp]
\centering
\includegraphics[width=\textwidth]{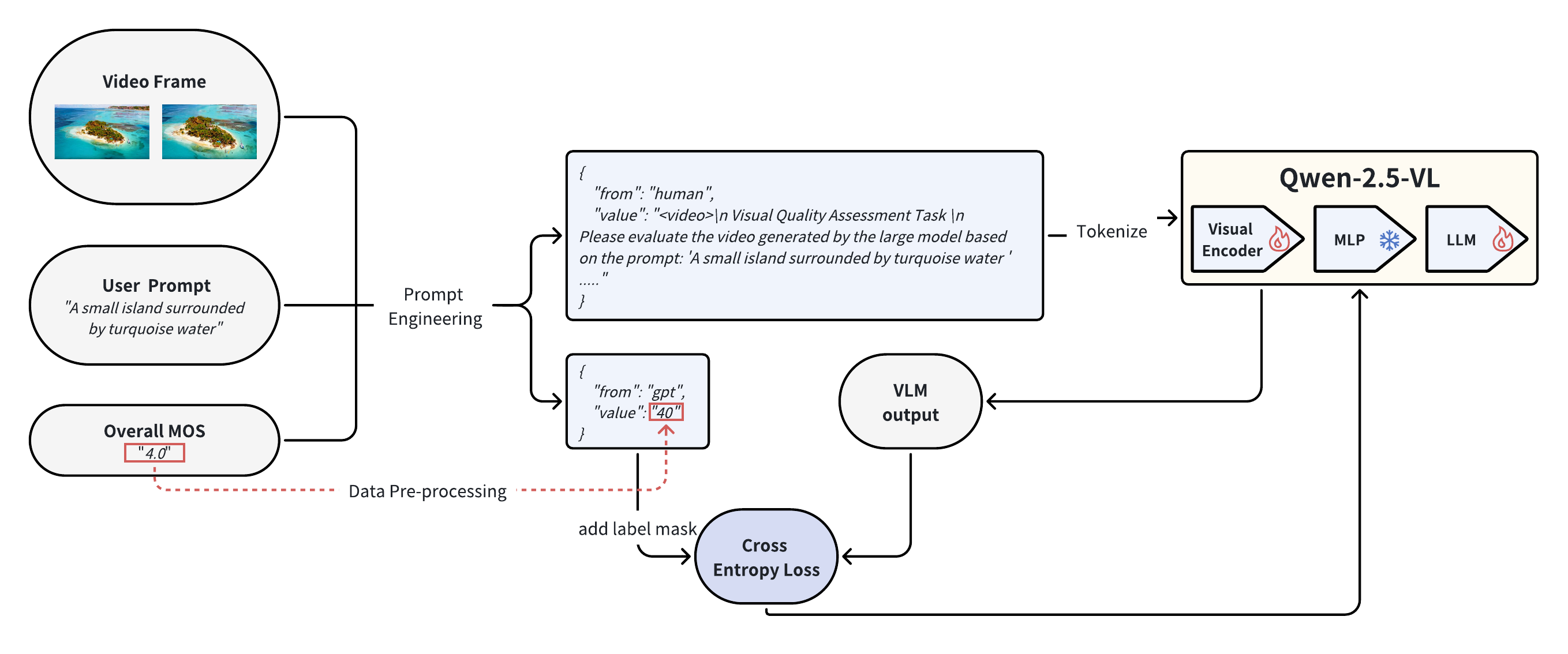}
\caption{Overall process of IOVQA}
\label{fig1}
\end{figure*}

\subsection{Text-to-Video Generative Models}
Recent developments in the field of diffusion models, for instance DDPM\cite{DDPM} and LDM\cite{HRIS}, have significantly accelerated the progress of Text-to-Video (T2V) generation technology. Given a text prompt, T2V generative models like VideoCrafter\cite{videocrafter}, StreamingT2V\cite{StreamingT2V} and Lumiere\cite{Lumiere} can synthesize entirely new videos not seen before. Early diffusion-based video models like LAVIE\cite{LAVIE} and SEINE\cite{SEINE} are typically built on top of Text-to-Image (T2I) models and are expanded into the video domain by adding a temporal module. Recent ones are trained from scratch directly on video data. Among them, MagicVideo\cite{MagicVideo} and Latent-Shift\cite{Latent} such models based on Latent Diffusion Models (LDMs) have received extensive attention for their effectiveness and efficiency. Additionally, Photorealistic\cite{Photorealistic} and GenTron\cite{GenTron} using pixel-based Diffusion Transformers (DiT) have also achieved high-quality results.

\subsection{Video Quality Assessment} 

Given the current development phase of text-to-video generation models, the distance to achieve their ultimate goal remains unclear, prompting researchers to develop evaluation methods for benchmarking these models. Common approaches involve video frame quality assessment and text frame alignment using metrics such as FVD\cite{fvd} and clip\cite{clip}, respectively. However, such metrics are not capable of capturing other key dimensions including video-theme consistency, temporal consistency, and factual accuracy and so on. Recent efforts like VBench\cite{VBench} have proposed metrics that utilize tools such as DINO\cite{DINO} and optical flow\cite{flow} to address these issues, yet their output exhibit low correlation with human perception. For example, most models score extremely high in background consistency on VBench\cite{VBench}, overestimating current T2V capabilities. Other methods including EvalCrafter\cite{Evalcrafter} rely on human raters for comprehensive evaluation.\par
Other recent studies adopt strategies of prompting multi-modal large language models like Gemini\cite{Gemini} and GPT-4o\cite{GPT4o} for video quality assessment, such as VideoPhy\cite{Videophy} and VIEScore\cite{Viescore} , though resulting in poor alignment with human evaluators. Meanwhile, T2VQA\cite{T2VQA} proposes training assessment models on human-annotated rating.

\subsection{VLM-based Video Quality Assessment}
Notably, video-language-model-based quality assessment approaches like DeQA-Score\cite{you2025teaching}, Q-align\cite{Q-align} and Q-insight\cite{Q-insight} have gained notable result by combining the reasoning capabilities of large language models with powerful score regression abilities. In VQA tasks, VQA-Scorer\cite{VQA-Scorer} integrated a SlowFast-R50 encoder to enhance motion capture and applied instruction tuning to multi-modal large language models to focus on low-level visual dimensions. In further tasks of evaluating videos generated by T2V models, demanding more complex fine-grained analysis, VideoScore\cite{Videoscore} realized automatic assessment by training a VLM on the large-scale human-annotated VideoFeedback dataset with variable aspects; VisionReward\cite{Visionreward} put forward a hierarchical visual assessment pipeline and multi-dimensional preference learning to fit fine-grained human preferences for image and video generation; and UnifiedReward\cite{Unified} introduced a unified preference learning framework supporting joint pairwise ranking and point-wise scoring for multi-modal generation and understanding.
\begin{figure*}[tbp]
\centering
\includegraphics[width=\textwidth]{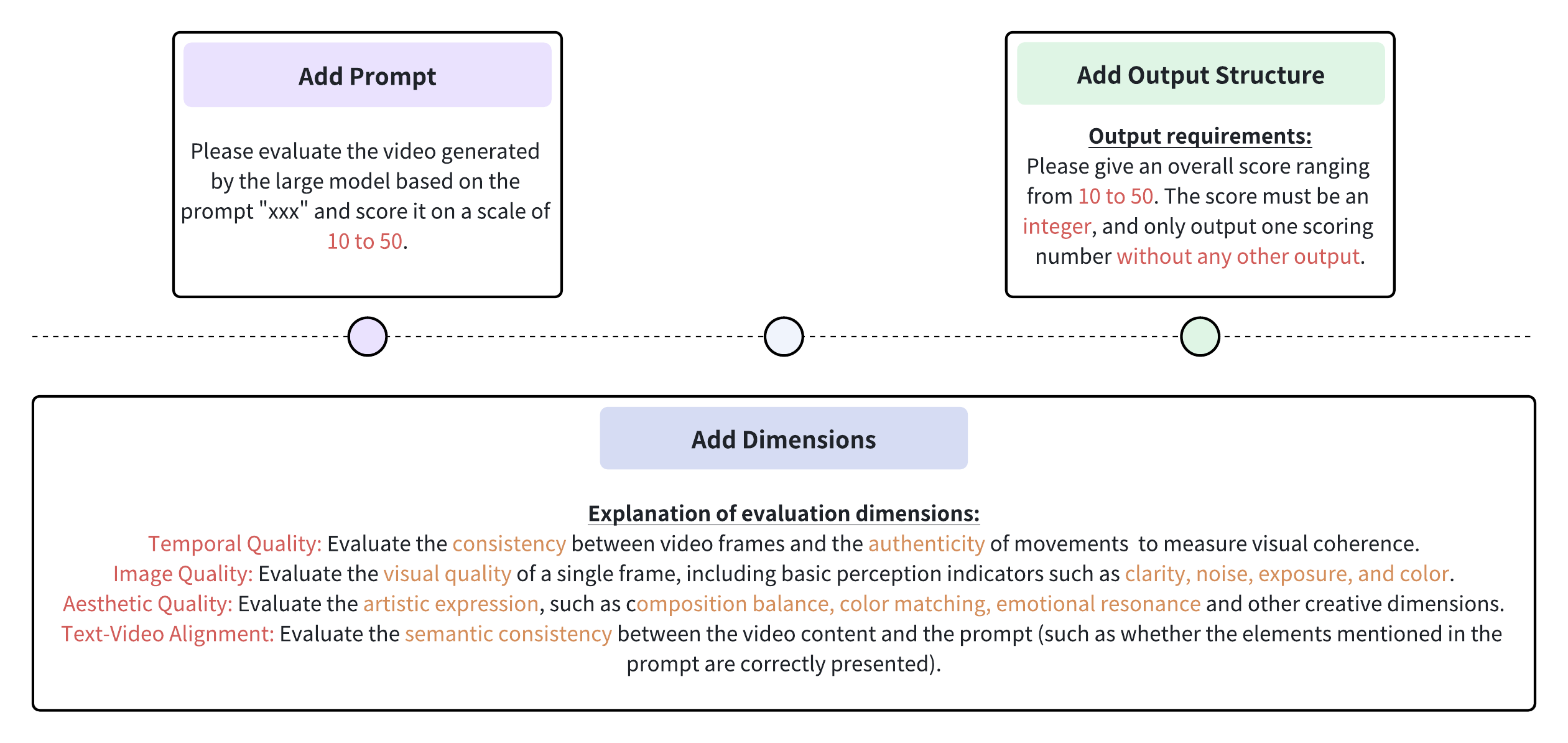}
\caption{Prompt engineering}
\label{fig2}
\end{figure*}

\subsection{Qwen2.5-VL}
Qwen2.5-VL\cite{qwen25} is the latest flagship model of the Qwen vision-language series, demonstrating significant advancements in both basic capabilities and innovative functions. We choose this VLM as the base model because of its great breakthroughs in enhanced visual recognition, precise object positioning and ability for structural output. 

\subsubsection{Model Architecture}
Large Language Model (LLM): The Qwen2.5-VL series uses the Qwen2.5 LLM as a base component and initializes it with the appropriate pre-trained weights. To better meet the needs of multi-modal understanding, the 1D rotational position embedding (1D RoPE) has been modified to a multi-modal rotational position embedding (MRoPE) adapted to absolute time.
Visual Encoder: The Qwen2.5-VL visual encoder uses a newly-structured vision transformer (ViT) architecture. This architecture introduces two-dimensional rotational position embedding (2D-RoPE) and a window focus mechanism to deal with native input resolution and accelerate the computation of the entire visual encoder. This visual encoder parts the image into patches with a stride of 14 to convert it into a set of image features. \par 
MLP-based Vision-Language Merger: It compresses the feature sequence dynamically before inputting it into the large-language model by grouping the four spatially adjacent patch features extracted by the vision transformer. Then it concatenates these grouped features, and process them through a two-layer MLP to project these features into a space consistent with the text embedding dimension used in the LLM.

\subsubsection{Key Features for VQA tasks}
There are two key features of Qwen2.5-VL that benefit in VQA tasks.\par
\textbf{Precise object grounding across formats: } 
This ability suggests a significant improvement in fine-grained spatial understanding, which enables the model to not only recognize and classify objects in visual content, but also accurately quantify their positions through absolute coordinates, bounding boxes, or polygon annotations, and even count instances with high precision. \par
In video generation or processing, artifacts including object truncation or blurring at critical boundaries can directly reduce perceptual quality. With the capability of locating objects with pixel level accuracy, the model can quantify these defects, such as measuring the percentage of bounding boxes falling outside the frame or the sharpness of edges within the localization area, transforming subjective defects into objective indicators, thereby improving the quality of video quality assessment.\par

In addition, for video-generative models, quality is often related to the alignment of the videos to text prompts, for instance, 'a cat sitting on a red sofa'. Accurate positioning enables the model to verify whether the space conforms to these prompts, that is, to confirm that the cat is indeed within the bounding box of the sofa, and that the positioning area of the sofa matches the color described in this example. This narrows the gap between semantic intent and visual execution, which is a key quality dimension that may be easily neglected by coarse-grained evaluation metrics.

\textbf{Ultra-long video understanding and fine-grained video grounding: }

This ability elevates the model's temporal inference beyond short clips, enabling it to process and interpret videos lasting hours while maintaining accuracy in locating key events or anomalies at the second level. By extending the dynamic resolution to the temporal dimension, the model adapts its processing granularity to the inherent rhythm of the video: it can perform detailed analysis on fast-paced segments and maintain efficiency on slower static segments, while preserving context throughout the entire timeline. \par

In VQA tasks, many quality issues manifest in brief, localized segments, like a 2-second blur during a critical action, a 3-second audio-visual de-synchronization in a dialogue, or a 5-second repetition of frames in a generative video. Fine-grained temporal localization allows the model to figure out and locate these short-time defects, instead of averaging them out across the entire video. For example, in the evaluation of a 2-hour generative video based on a prompt, the model is capable of figuring out a 2-second segment that fails to render a required event like "a door opening", even if the rest of the video is coherent. This ability benefits in capturing the kind of "critical moment failure" that disproportionately deduces quality in human perception.

\section{Method}

Supervised fine-tuning (SFT) is a key stage in deploying large language models (LLMs) and multimodal large language models (MLLMs). In this process, a pre-trained foundation model is further optimized on a task-specific, human-annotated dataset to generate response with targeted behaviors, domain-specific knowledge, or judgments closer to human perception. Different from pre-training aiming at learning general patterns from large amounts of unlabeled data, SFT focus on specific tasks including quality assessment, question answering, or content generation by exposing it to prompt-label pairs that explicitly demonstrate the correct mapping, for instance, pairing a video clip with a overall MOS from human based on four dimensions in this challenge, or pairing a text prompt with a contextually appropriate response. \par

This process expand the knowledge base of the model’s parameter through supervised gradient descent of the loss function and cross-entropy in our implementation. So it can perform predictions beyond the pre-training distribution and generate outputs that are not only grammatically coherent but also semantically consistent with the task objectives. For VLMs including Qwen, SFT enhances the integration of visual and textual modalities, enabling the model to better interpret cross-modal relationships, such as linking video frames to evaluation criteria in prompts in this challenge, and generate consistent, task-relevant outputs. By narrowing the gap between general pre-training and specific application requirements, SFT does great benefit in mimicing human perception, thus enhancing model's reliability, reducing hallucinations, and ensuring the meet of accuracy and alignment requirements of downstream academic, industrial, or real-world use cases. \par

Our SFT process is illustrated in Fig\ref{fig1}.

\subsection{Data Pre-processing}
The dataset used for fine-tuning the large model consists of three core components: video, user prompt, and overall\_MOS scores (a decimal value ranging from 1 to 5). To ensure that the dataset for fine-tuning aligns with the task objectives and benefits for effective training, a series of pre-processing steps are implemented. The detailed operations and their underlying rationales are elaborated as follows:

\subsubsection{Video Frame Sampling}
Since the video content exhibits minimal semantic variation throughout its duration, with only subtle changes in perspective and motion, extracting all frames would introduce redundant visual data, increasing computational complexity during model training without contributing meaningful information. So we extract 1-2 frames each video, retaining the essential visual features required for evaluation while reducing data volume effectively.\par

Given the limited variation in the video, 1-2 frames are sufficient to capture the core visual content. For instance, if the video depicts a static scene with slight camera panning, a single frame from the middle section can adequately represent the overall visual characteristics. In cases where there are minor dynamic changes like slow movement of objects, two frames can capture the range of visual variations, ensuring the sampled data is representative of the entire video. Considering the evaluation metrics(aesthetic quality, image quality, temporal quality, and text-video alignment), only temporal quality takes the difference between similar frames into consideration, 2 frames may be better but 1 is enough for limited computation resource.\par

Moreover, large models, especially multi-modal models that process both visual and textual data, face challenges in handling high-dimensional video inputs. Reducing the number of frames to 1-2 significantly lowers the dimensionality of the visual input, enabling the model to focus on integrating visual information with textual prompts more efficiently. This optimization accelerates the training process and improves efficiency in model input handling.

\subsubsection{Prompt Engineering}
The user prompt, which reflects the user's requirements for the video, is embedded into a new prompt for the evaluation model. This transformation is guided by the following principles:\par
\textbf{Clarification of Evaluation Dimensions: }
The original user prompt specifically expresses the user's expectations for generated video. By incorporating the user prompt into the a prompt and explicitly stating the evaluation dimensions (aesthetic quality, image quality, temporal quality, and text-video alignment), we provide the model with a clear framework for assessment. This ensures that the model's evaluation is aligned with the user's fine-grained requirements, avoiding deviations caused by ambiguous instructions.\par
\textbf{Standardization of Output Format: }
We restricting the model to output only a single number without any explanatory text, which is critical for a more direct fine-tuning process, ensuring the consistency and usability of the results. In downstream tasks such as model training and performance evaluation, a standardized numerical output simplifies data processing and comparison. Without this constraint, the model might generate lengthy explanations, which would require additional parsing and could introduce noise into the training process.\par
The final input prompt is illustrated in Fig\ref{fig2}.

\subsubsection{Transformation of Overall\_MOS into Label}

\begin{figure}[tbp]
    \centering
    \includegraphics[width=\linewidth]{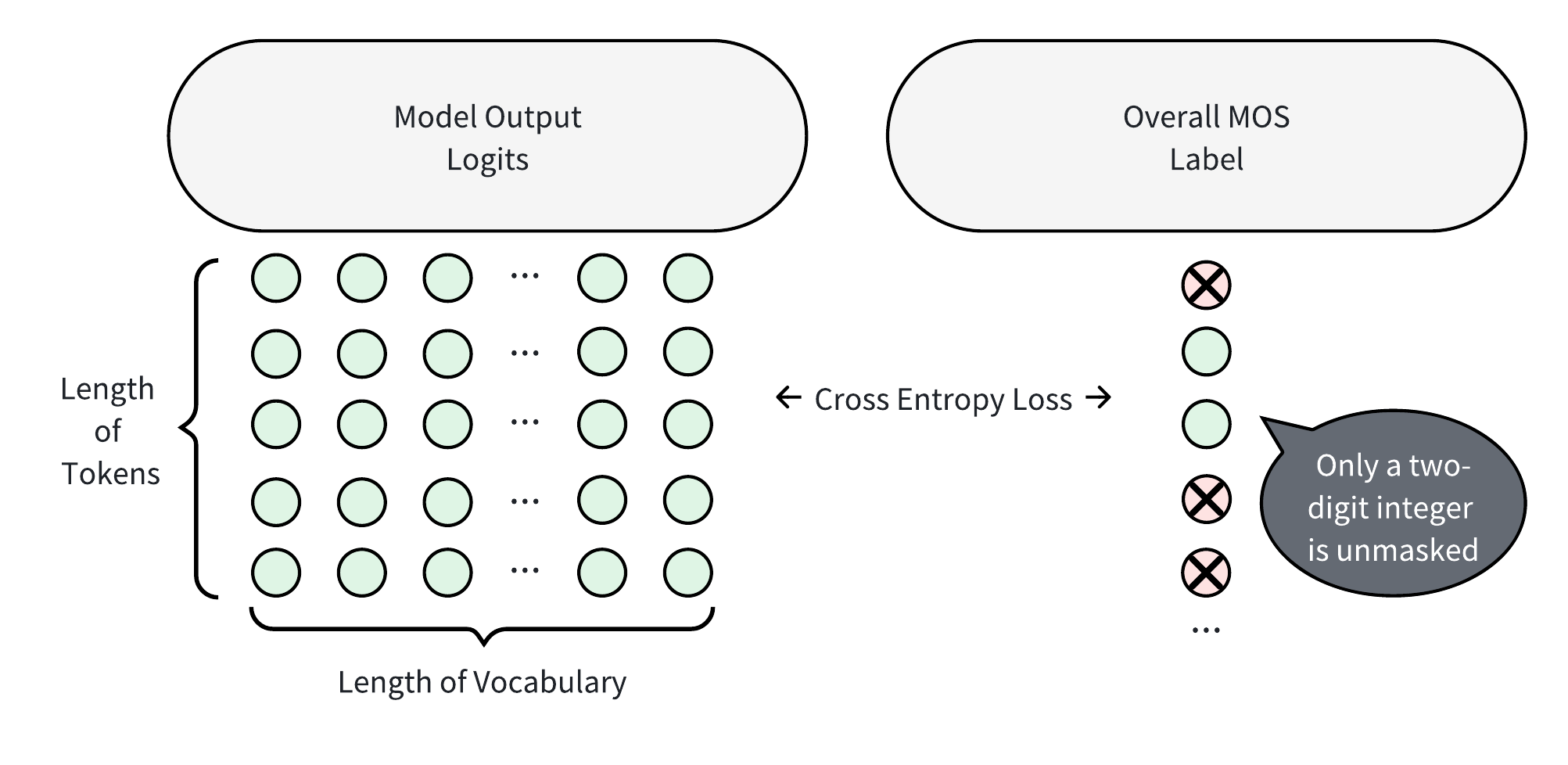}
    \caption{Integer-Only Mask}
    \label{fig3}
\end{figure}

The Overall\_MOS score is processed by retaining one decimal place, multiplying by 10, and converting it into an integer within the range of [10,50], serving as the label for the evaluation model's output. The reasons for this transformation are as follows:\par
\textbf{Reduction of Label Granularity: }The original Overall\_MOS is a decimal with potentially high precision (3.666), since human judgments of quality are averaged between 15 ones, operating at a coarser granularity. Retaining one decimal place (rounding 3.666 to 3.7) simplifies the label while still preserving the essential information about the quality level, and there is no need for more decimal places because the rating of human is limited to only 5 levels.\par
\textbf{Scaling to a Suitable Range:} We convert the rounded Overall\_MOS to an integer between 10 and 50 (by multiplying by 10). This transformation expand the original range [1, 5] into a wider range [10, 50], which provides a larger margin for distinguishing between different quality levels. This increased range enhances the model's ability to learn fine-grained differences in quality during training. Additionally, integer labels utilize the inherent advantage of LLMs in predicting integer values over decimal fractions, rooted in their auto-regressive nature as categorical predictors. This compatibility improves the efficiency and stability of the training process.\par
\textbf{Consistency with Prompt Constraints:} We ensure that the model's training targets are consistent with the expected output format. This alignment is crucial for the model to learn the mapping between inputs (visual frames and prompts) and outputs (labels), as it creates a direct correspondence between the training data and the task requirements.\par
\textbf{Facilitation of Loss Calculation: }During model training, the loss function compares the model's predicted output with the true label. We propose a label masking strategy illustrated in Fig\ref{fig3} to refine the training process of integer score prediction, where a mask is applied to the labels such that only two-digit integer scores are retained for cross-entropy computation. Specifically, during the loss calculation phase, irrelevant components in the original labels like explanatory text or redundant tokens are masked out, leaving only the integer score as the target for the model to learn.

\subsection{Softmax regression}
\begin{figure}[tbp]
    \centering
    \includegraphics[width=\linewidth]{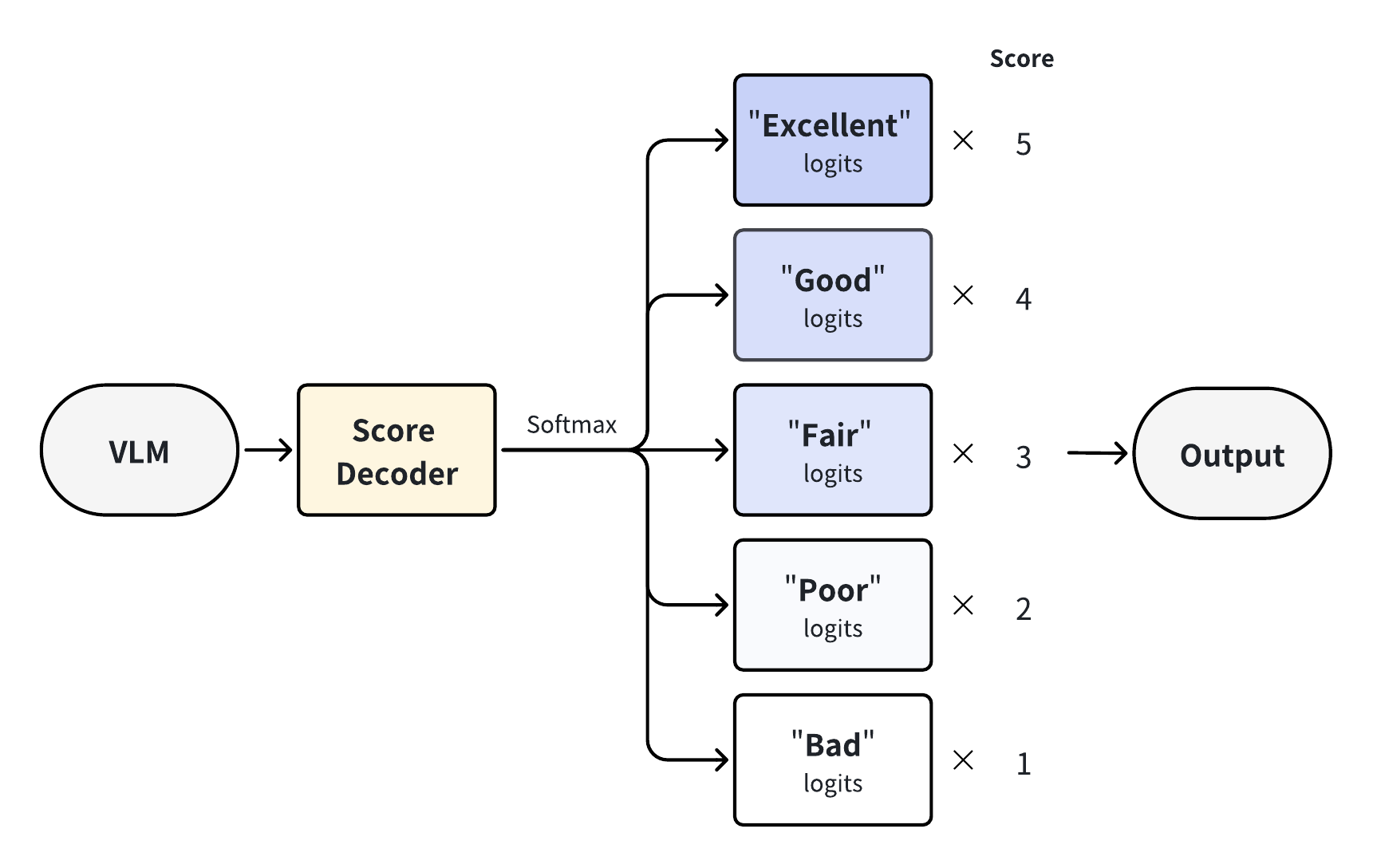}
    \caption{Soft regression Method for Ensembling}
    \label{fig3}
\end{figure}
We also come up with a similar way combining integer labels with softmax regression, shown in Fig. This method define video quality into five grades: bad, poor, fair, good, and excellent, corresponding scores of 1 to 5 and add a special token $\left\langle LABEL\_1\right \rangle$ to the vocabulary to capture video quality representations. It then extracts the hidden state of this special token as the video quality representation and obtains 5 implicit logits by passing the hidden state through an MLP. These logits are then converted into probabilities for the five grades via softmax, and the predicted score is derived through weighted summation.\par
The essential difference between two methods is that IOVQA converts decimal scores into 40 categories, specifically an integer between 10 and 50, while the other method still utilizes grades from 1 to 5 and synthesize decimals by weighted summation based on probability.

\subsection{Ensembling}
Whether it is PLCC or SRCC, their key requirements are the accuracy and robustness of the prediction results. The ensembling strategy serves both because of its reduction in overfitting and capability of complementary fusion. A single model is prone to overfit the noise in the training data, resulting in a decrease in PLCC/SRCC on the test set. By combining multiple finetuned VLMs with various parameter counts, the integration reduces the risk of overfitting and makes the prediction more in line with the true pattern of the data.\par
Moreover, different models have different levels of abilities in feature extraction and connection of visual and test features, along with different error distribution. The integration can synthesize these differences, thus covering a more comprehensive range of prediction scenarios and reducing the blind spots of a single model.\par
As a result, we ensemble five models, including two 7B Qwen2.5-VL models fine-tuned with different lora\_r settings, a 32B Qwen2.5-VL model, a 72B Qwen2.5-VL model and a 7B Qwen2-VL model using the softmax regression strategy explained above. The weight of each model is set to 0.25, 0.15, 0.25, 0.1 and 0.25.

\section{Experiment}
\subsection{Dataset}
\begin{figure}[tbp]
    \centering
    \includegraphics[width=\linewidth]{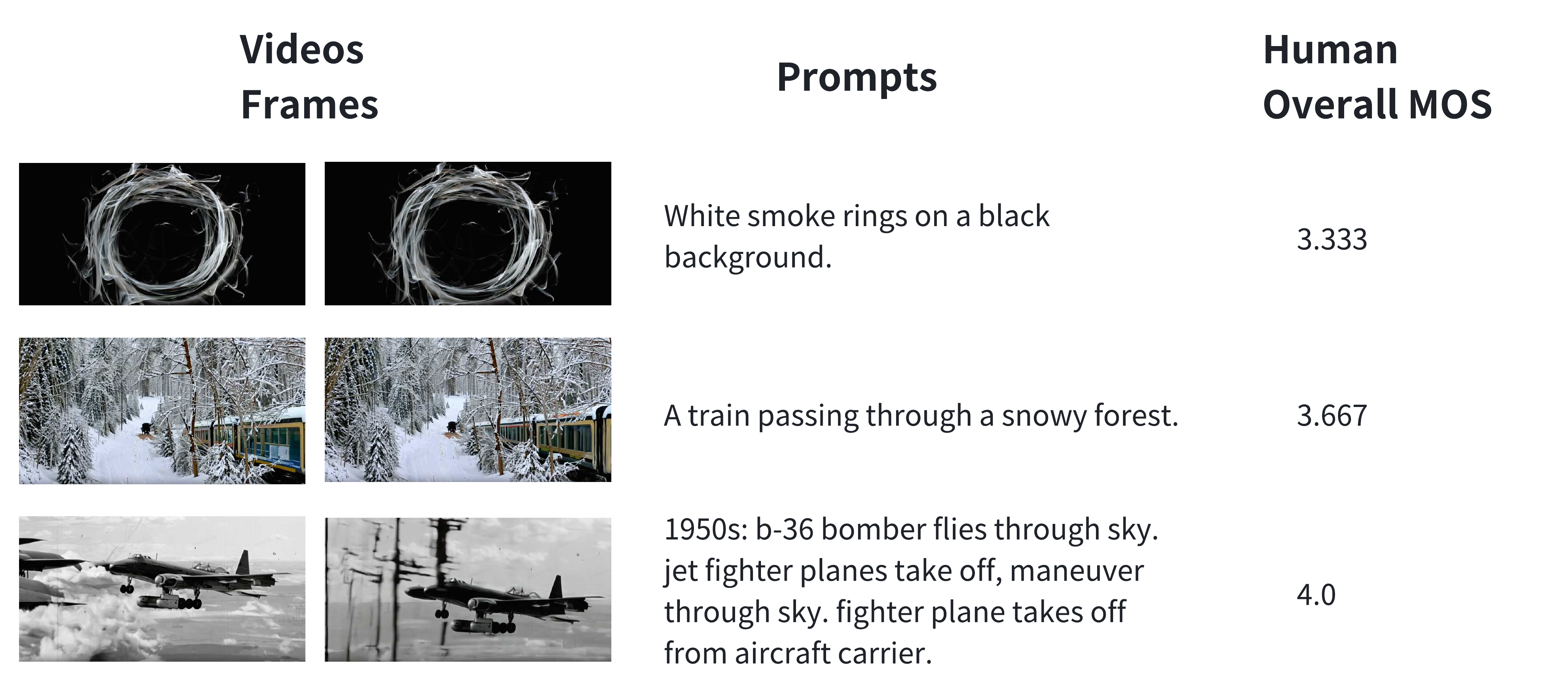}
    \caption{Examples of dataset}
    \label{fig4}
\end{figure}
Our experimental studies employ the Taobao Video Dataset - Generated Content (TaobaoVD-GC), designed for multi-dimensional evaluation of AIGC videos. It consists of 5000 AIGC videos generated from 5 representative Text-to-Video (T2V) generative algorithms, and each video-text pair is annotated by 15 participants' subjective scores after anomaly filtering. The duration and resolutions of the videos are 4 to 5 seconds, and 720 × 1280 or 768 × 1360, respectively.\par
The dataset is divided into a training set, a validation set, and a test set.
The training set contains 4,000 video-text pairs for
model parameter learning; the validation set and the test set each comprises approximately 500 pairs for performance evaluation.

\subsection{Models}
For IOVQA, we fine-tuned 3 pre-trained models: Qwen2.5-VL-7B-Instruct, Qwen2.5-VL-32B-Instruct and Qwen2.5-VL-72B-Instruct.\par

\subsection{Training Parameters}
For IOVQA, we fine-tuned 4 models using 8 NVIDIA-H20 GPUs, the details are listed in Tab\ref{tab:train}.\par
\begin{table}[htbp]
    \centering
    \begin{tabular}{c|cccc}
        \toprule
        Model ID & 1 & 2 & 3 & 4\\
        \hline\hline 
        Model size & 7B & 7B & 32B & 72B \\
        Lora\_r & 32 & 128 & 128 & 128\\
        Lora\_alpha & 32 & 32 & 32 & 32\\
        Lora\_dropout & 0.1 & 0.1 & 0.1 & 0.1\\
        Training epoch & 6 & 6 & 6 & 6\\
        Used epoch  & 2 & 3 & 4 & 5 \\
        \bottomrule
    \end{tabular}
    \caption{Details of used fine-tuned models}
    \label{tab:train}
\end{table}
Other parameter settings used are listed in Tab\ref{tab:para}
\begin{table}[htbp]
    \centering
    \begin{tabular}{c|c}
        \toprule
        Parameter & Setting\\
        \hline\hline 
        Learning rate & 3e-4\\
        Weight decay & 1e-4\\
        Warmup ratio & 0.1\\
        Lr scheduler type & cosine\\
        Optimizer & Adamw\_torch\\
        Adam epsilon & 1e-8\\
        Adam beta1 & 0.9\\
        Adam beta2 & 0.98\\
        DeepSpeed & zero3\\
      
        \bottomrule
    \end{tabular}
    \caption{Training parameter settings}
    \label{tab:para}
\end{table}

\subsection{Evaluation Metrics}
To comprehensively assess model performance, the Spearman
Rank Correlation Coefficient (SRCC) and Pearson Linear Correlation Coefficient (PLCC) are used as evaluation metrics. \par
Spearman Rank Correlation Coefficient (SRCC) and Pearson Linear Correlation Coefficient (PLCC), defined as \cref{eq:srcc} and~\cref{eq:plcc}, are commonly employed as evaluation metrics for quality assessment. They measure the ranking ability and fitting ability of a prediction model respectively.\vspace{1em}\begin{equation}	SRCC=1- \frac{6\sum_{m=1}^{M}r_{m}^{2}}{M\left ( M^{2}-1 \right )}	\label{eq:srcc}\end{equation}\begin{equation}    PLCC\!=\!\frac{\sum_{m=1}^{M}\!\left ( Q_{m}-\bar{Q} \right )\left ( Q'_{m}-\bar{Q'} \right )}{\sqrt{\sum_{m=1}^{M}\!\left ( Q_{m}\!-\!\bar{s} \right )^2}\sqrt{\sum_{m=1}^{M}\!\left ( Q'_{m}\!-\!\bar{Q'} \right )^2}}	\label{eq:plcc}\end{equation}where $r_{m}$ is the rank difference of the subjective quality score $Q_{m}$ and the predicted score $Q'_{m}$. \par
The final result is the average of both score. 
\begin{equation}
Final Score = (SRCC_{overall} + PLCC_{overall}) / 2
\end{equation}

\subsection{Experimental Results}
We validated the effectiveness of our method on the test dataset of VQualA 2025 GenAI-Bench AIGC Video Quality Assessment Challenge -- Track I and the result ranked 3rd among all the methods on leaderboard.

\begin{table}[htbp]
    \centering
    \begin{tabular}{c|ccc}
        \toprule
        User & Final score & SRCC & PLCC\\
        \hline\hline 
         MLLM-VQA & 0.72 & 0.71 & 0.72\\
        SJTU-IntMeGroup & 0.71 & 0.71 & 0.71\\
        \textbf{Ours} & 0.70 & 0.70 & 0.70\\
         MM & 0.69 & 0.68 & 0.70\\
         QA-Veteran & 0.69 & 0.69 & 0.69\\
          no-no & 0.66 & 0.66 & 0.66\\
          Ikshana & 0.57 & 0.56 & 0.58\\
           Gzu-Team & 0.56 & 0.55 & 0.58\\
        \bottomrule
    \end{tabular}
    \caption{Comparison with results of the VQualA 2025 GenAI-Bench AIGC Video Quality Assessment Challenge -- Track I}
    \label{tab:para}
\end{table}

We also compared IOVQA with a similar way combining integer labels with softmax regression. This method define video quality into five grades: bad, poor, fair, good, and excellent, corresponding scores of 1 to 5 and add a special token $\left\langle LABEL\_1\right\rangle$ to the vocabulary to capture video quality representations. It then extracts the hidden state of this special token as the video quality representation and obtains 5 implicit logits by passing the hidden state through an MLP. These logits are then converted into probabilities for the five grades via softmax, and the predicted score is derived through weighted summation.\par
The essential difference between two methods is that IOVQA converts decimal scores into 40 categories, specifically an integer between 10 and 50, while the other method still utilizes grades from 1 to 5 and synthesize decimals by weighted summation based on probability.

\begin{table}[htbp]
    \centering
    \begin{tabular}{c|c|c}
        \toprule
        Method & Validation dataset & Test dataset\\
        \hline\hline 
        Softmax regression  & 0.55 & 0.60\\
        IOVQA & 0.60 & 0.64\\
        \bottomrule
    \end{tabular}
    \caption{Comparison with softmax regression method}
    \label{tab:para1}
\end{table}

\subsection{Ablation Study}
\subsubsection{Integer-Only Mask}
In Tab\ref{tab:abla}, we compare the result of different supervised fine-tuning methods to demonstrate the effectiveness of integer-only mask. The baseline result is Qwen2.5-VL-7B-Instruct without any fine-tuning, and the other three methods are fine-tuning the same base model using decimal labels, using integer labels without integer-only mask and using integer labels with integer-only mask. The comparison demonstrates that using integer labels outperforms decimal labels by 0.03, confirming that integer-formatted targets align better with VLMs' inherent strengths in discrete value prediction. \par   
Additionally, adding the integer-only mask to integer labels further boosts the score to 0.64, which validates that limiting loss computation to responses containing merely integer scores reduces noise from irrelevant components, thereby enhancing prediction accuracy. Both the integer label conversion and the integer-only mask contribute synergistically to IOVQA's performance.\par

\begin{table}[htbp]
    \centering
    \begin{tabular}{c|c}
        \toprule
        Method & Final score\\
        \hline\hline 
        Not fine-tuned & 0.18 \\
        Decimal labels & 0.60\\
        Integer labels & 0.63\\
        Integer labels and integer-only mask & \textbf{0.64}\\
        \bottomrule
    \end{tabular}
    \caption{Ablation study of fine-tuning methods}
    \label{tab:abla}
\end{table}

\subsubsection{Baseline Models}
In Tab\ref{tab:abla1}, we compare the best results of Qwen-2.5-VL of different sizes. All of the fine-tuning process are with same parameter settings except for lora\_r. \par

\begin{table}[htbp]
    \centering
    \begin{tabular}{c|ccc}
        \toprule
        Model size & Lora\_r & Final score & Best epoch\\
        \hline\hline 
        7B & 32 & 0.614 & 2\\
        7B & 128 & 0.642 & 3\\
        32B & 128 & 0.640 & 4\\
        72B & 128 & 0.622 & 5\\
        \bottomrule
    \end{tabular}
    \caption{Ablation study of different model size and lora\_r}
    \label{tab:abla1}
\end{table}
The results suggest that for the 7B model, increasing lora\_r from 32 to 128 leads to a noticeable improvement in the final score (from 0.614 to 0.642), indicating that a larger LoRA rank allows the model to capture more task-specific features in video quality assessment. But when scaling up to larger models (32B and 72B) with lora\_r fixed to 128, the performance does not follow a monotonic upward trend. This suggests that simply increasing model size may not guarantee better performance in this specific task, the reason of which may lies in limited size of dataset, leading to overfitting or mismatches between model capacity and task complexity. The 7B model with LoRA rank of 128 emerges as the optimal choice, balancing efficiency and performance for video quality assessment under our experimental setup.
\section{Conclusion}
In this paper, we propose IOVQA, a novel method that integrates label construction with a targeted loss calculation mechanism. Specifically, we first convert human-assigned Mean Opinion Score (MOS) labels, originally represented as decimal values, into integer labels. During the loss computation process, an integer-only masking strategy is employed, focusing exclusively on the loss derived from integer scores rather than incorporating the full range of sentence outputs.This design leverages VLMs' superior ability to predict integer values compared to decimal numbers. By aligning the label format and loss computation with VLMs’ natural capabilities, IOVQA enables more accurate and robust video quality assessment.\par
Experimental results on the Taobao Video Dataset - Generated Content validate the effectiveness of our approach, demonstrating that it enhances VLMs’ performance in targeted video quality assessment tasks while bringing their evaluations closer to human perceptual judgment.
{
    \small
    \bibliographystyle{ieeenat_fullname}
    \bibliography{main}
}

\end{document}